\documentclass[sigconf]{acmart}
\AtBeginDocument{%
  \providecommand\BibTeX{{%
    \normalfont B\kern-0.5em{\scshape i\kern-0.25em b}\kern-0.8em\TeX}}}

\copyrightyear{2023} 
\acmYear{2023} 
\setcopyright{acmlicensed}\acmConference[MM '23]{Proceedings of the 31st ACM International Conference on Multimedia}{October 29-November 3, 2023}{Ottawa, ON, Canada}
\acmBooktitle{Proceedings of the 31st ACM International Conference on Multimedia (MM '23), October 29-November 3, 2023, Ottawa, ON, Canada}
\acmPrice{15.00}
\acmDOI{10.1145/3581783.3612855}
\acmISBN{979-8-4007-0108-5/23/10}

\usepackage{adjustbox}
\usepackage{balance}
\usepackage{hyperref}

\begin{document}

\title{ Effect of Attention and Self-Supervised Speech Embeddings on Non-Semantic Speech Tasks}

%
% The "author" command and its associated commands are used to define
% the authors and their affiliations.
% Of note is the shared affiliation of the first two authors, and the
% "authornote" and "authornotemark" commands
% used to denote shared contribution to the research.
\author{Payal Mohapatra}
\authornotemark[1]
\affiliation{%
  \institution{Northwestern University}
  % \streetaddress{P.O. Box 1212}
  \city{Evanston}
  \state{Illinois}
  \country{USA}
  \postcode{60202}
}
\email{payal.mohapatra@northwestern.edu}

\author{Akash Pandey}
\authornotemark[1]
\affiliation{%
  \institution{Northwestern University}
  % \streetaddress{P.O. Box 1212}
  \city{Evanston}
  \state{Illinois}
  \country{USA}
  \postcode{60202}
}
\email{akash.pandey@northwestern.edu}

\author{Yueyuan Sui}
\affiliation{%
  \institution{Northwestern University}
  % \streetaddress{P.O. Box 1212}
  \city{Evanston}
  \state{Illinois}
  \country{USA}
  \postcode{60202}
}
\email{yueyuansui2024@u.northwestern.edu}

\authornote{All authors contributed equally to this research. Payal worked on the large language model embeddings for feature extraction, Akash worked on the attention module,  and Yueyuan worked on the data preparation and model selection.}


% \affiliation{%
%   \institution{Northwestern University}
%   % \streetaddress{P.O. Box 1212}
%   \city{Evanston}
%   \state{Illinois}
%   \postcode{60202}
% }

\author{Qi Zhu}
\affiliation{%
  \institution{Northwestern University}
  % \streetaddress{P.O. Box 1212}
  \city{Evanston}
  \state{Illinois}
  \country{USA}
  \postcode{60202}
}
\email{qzhu@northwestern.edu}


\begin{abstract}
Human emotion understanding is pivotal in making conversational technology mainstream. We view speech emotion understanding as a perception task which is a more realistic setting. With varying contexts (languages, demographics etc.) different \textit{share} of people perceive the same speech segment as a non-unanimous emotion. As part of the ACM Multimedia 2023 Computational Paralinguistics ChallengE (ComParE) in the \textbf{EMotion Share} track, we leverage their rich dataset of multilingual speakers and multi-label regression target of 'emotion share' or perception of that emotion. We demonstrate that the training scheme of different foundation models dictates their effectiveness for tasks beyond speech recognition, especially for non-semantic speech tasks like emotion understanding.  This is a very complex task due to multilingual speakers, variability in the target labels, and inherent imbalance in the regression dataset. Our results show that HuBERT-Large with a self-attention-based light-weight sequence model provides 4.6\% improvement over the reported baseline. 
\end{abstract}

\begin{CCSXML}
<ccs2012>
   <concept>
       <concept_id>10010147.10010178.10010179</concept_id>
       <concept_desc>Computing methodologies~Natural language processing</concept_desc>
       <concept_significance>500</concept_significance>
       </concept>
 </ccs2012>
\end{CCSXML}

\ccsdesc[500]{Computing methodologies~Natural language processing}

\keywords{Emotion share, large-language model, attention}


\maketitle

\section{Introduction}
% Motivate your problem
Human emotion understanding is vital to engage in meaningful social interactions. With the rising use of voice-assisted technology and the use of natural language as a key Human-Machine interface, it is important that we develop techniques to understand the affective aspects of speech~\cite{triantafyllopoulos2023overview}. We need to correctly identify and characterize the tone of speech and not merely understand its semantics of 'what' has been said. It is crucial to understand 'how' it has been said. Classifying a speech segment as one type of emotion is oversimplifying the way humans perceive language~\cite{cowen2021semantic}. On many occasions under various contexts, different groups of people perceive the same emotion differently. This gives rise to a viewpoint that we can assign a \textit{perceivability share} for an emotion in speech.

% What are the past works and their drawbacks

Traditionally, speech features like mel-frequency cepstral coefficients (MFCCs)~\cite{likitha2017speech, lalitha2015emotion}, filterbanks, fundamental frequency, energy, zero-crossing rate, chroma-based features, and their feature functionals~\cite{kakouros2023speech, aouani2020speech} are used with a rule-based or neural network classifier to recognize the emotions in speech. In recent times, tremendous progress of deep-learning models in the field of Natural Language Processing has positively boosted the performance of paralinguistic speech tasks~\cite{harar2017speech, cai2021speech, jahangir2021deep} like emotion recognition as well. Several works in the literature advocate the use of self-supervised pipelines (SSL)~\cite{wagner2023dawn, morais2022speech, yang2023ensemble} for Automatic-Speech-Recognition tasks for SER. The two foundation models that emerge as the current state-of-the-art feature extractors for ASR tasks are HuBERT~\cite{hsu2021HuBERT} and wav2vec2.0~\cite{baevski2020wav2vec}. 

Most if not all the previous works on understanding human emotions from speech can be categorized as Speech Emotion Recognition (SER). Such studies suffer from a few limitations - 1) the datasets consist of actors emulating certain emotions in controlled settings without much noise/interference, 2) focused on English speaking demographic and 3) hard labels for a speech segment~\cite{busso2008iemocap}. In this work, we move away from the emotion recognition formulation and instead focus on modeling a more complex task of the ratio of people perceiving the given speech for every emotion. We utilize the first-ever speech-emotion corpus containing the fraction of the annotator population who categorize it as each of the nine emotions - Anger, Boredom, Calmness, Concentration, Determination, Excitement, Interest, Sadness, and Tiredness. This dataset is released as part of the  Computational Paralinguistics Challenge (ComParE) 2023 \cite{baseline}. It contains data from multi-lingual speakers in real-world settings making the task more challenging. We want to design a robust architecture that can predict the 'share' of people who can perceive a given speech segment for each type of emotion. We provide a detailed analysis of the impact of different SSL embeddings on non-semantic downstream tasks like emotion perception. Non-semantic tasks rely on speech beyond its lexical meaning like language identification, speaker identification, emotion-related tasks, etc. We also discuss various architectures using custom attention, temporal convolution layers, long-short-term memory(LSTMs) and transformers, to learn a regressor for learning a continuous target for each emotion and provide insights.

The remainder of this paper is organized as follows. Section~\ref{sec:exp} describes the data preparation, the preliminaries on SSL models, and the implementation of various regressor architectures. Section~\ref{sec:results} discusses the various evaluation settings that allow us to discern our findings and present them in a comprehensive manner. Finally, Section~\ref{sec:conclude} summarises our contributions and discusses our future directions.
%% SSL + Emotion
% 1. https://ieeexplore.ieee.org/stamp/stamp.jsp?tp=&arnumber=9747870
% 2. Read this for the complexity of Emotion related tasks -- http://www.scholarpedia.org/article/Speech_emotion_analysis

% What do you propose?

% Tell about your paper organization

\section{Experimental Set up}\label{sec:exp}
We obtain the data from Hume AI as part of the Emotion Share Sub-Challenge of ACM Multimedia 2023 COMputational PARalinguistics challengE (ComParE) \cite{baseline}. More details on the data collection and statistics can be found in the baseline paper. This is a multi-lingual dataset with a wide demographic of participants. 9 different emotions are considered for each sample and are given a 'share' (based on the number of evaluators who rated the sample as a corresponding emotion) as the target \cite{cowen2019mapping}. This is a multi-task regression where we want to predict a continuous target for each sample for every emotion.
\begin{figure}[h]
  \centering
  \includegraphics[width=100mm, scale=1]{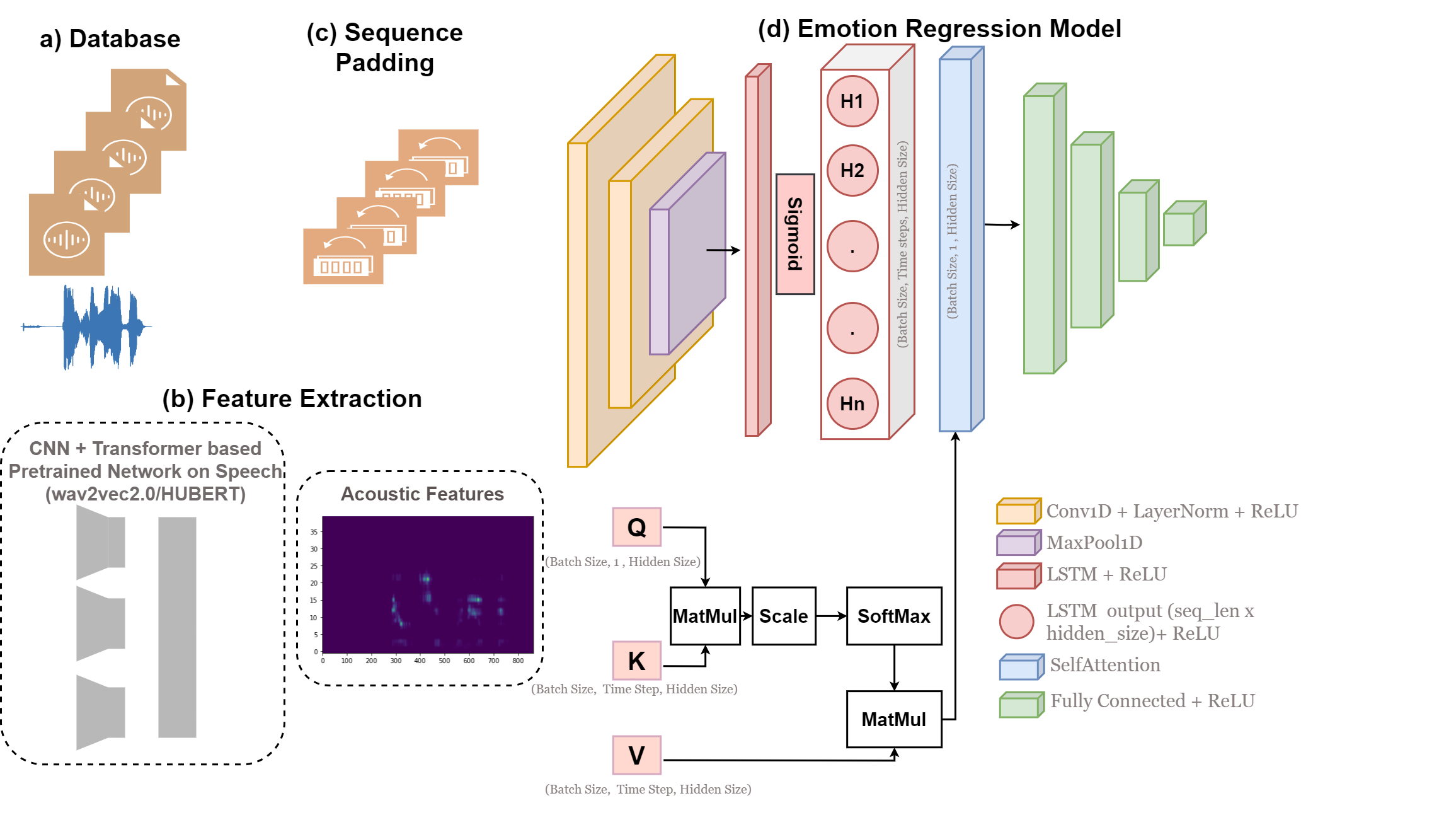}
  \caption{Overall architecture of our approach. a) Accessing the .wav files, b) Extracting embeddings from pretrained models or using acoustic features, c) Preparing dataset for training/evaluation by zero-padding the extracted features to deal with sequences of varying length, d) Using embedding/acoustic features for emotion share prediction. }
 \label{fig:Architecture_Diagram_}
\end{figure}
\vspace{-6mm}
\subsection{Data Preparation } 
The audio files vary in size from 1s to 8s. To support different audio lengths we use the technique of padding and masking in the following sequence models. Instead of masking the raw audio-files to the length of the longest audio segment, we conduct feature-padding as shown in Fig. \ref{fig:Architecture_Diagram_}.(c). After we extract the features using pretrained models on speech and acoustic features, we pad zeros to the temporal axis for each sample to match the maximum sequence length. In the case of wav2vec2.0 and HuBERT embeddings (both BASE and LARGE) we have a maximum sequence length of 398. For acoustic features, our maximum sequence length is 797. We store a dictionary of all the original sequence lengths as well, so that deeper in the architecture when operating on representations, we only consider the non-padded features. 

% \subsection{Feature Extraction}

% 
\vspace{-2.5mm}
\subsection{Pretrained Self-Supervised Speech Features} 
% % Refer 
% 1. https://dl.acm.org/doi/pdf/10.1145/3503161.3551606
% 2. https://ieeexplore.ieee.org/stamp/stamp.jsp?tp=&arnumber=10089511&tag=1
% 3. https://www.overleaf.com/project/6408c47ad2f55c1281ba52c8
% 4. https://jonathanbgn.com/2021/10/30/HuBERT-visually-explained.html

Transformer-based large language models are the most sought after techniques to derive feature embeddings for a given speech segment. In this work, we explore two state-of-the-art self-supervised pipelines for feature extraction, wav2vec2.0~\cite{baevski2020wav2vec} and HuBERT~\cite{hsu2021HuBERT}. The central architecture of both these models looks very similar superficially, consisting of convolution layers to extract latent feature embeddings from the raw audio followed by transformer layers. However, their training procedures are completely different. wav2vec2.0 uses contrastive loss by masking the latent feature embeddings and measuring the similarity between the predicted latent embedding versus the original feature embedding. They also leverage techniques like quantization and diversity loss to ensure robustness and avoid learning oversimplified representations. HuBERT on the other hand clusters the audio segments based on their latent feature embeddings using the K-Means algorithm and optimizes a cross-entropy loss. It chases the idea of discovering hidden units in a language rather than obtaining granular word-level representations as in the case of wav2vec2.0.

Both of these models are available in two sizes - BASE (~90M parameter) and LARGE (~300M parameters), with embedding sizes of 768 and 1024 respectively. Most previous works on emotion tasks are formulated as recognition. In this study, we address the unique task of predicting a continuous emotion perception target for a given audio segment for 9 different emotions. This is a more complex task than the former.

\begin{table*}[htbp]
\begin{adjustbox}{width=150mm,center}
\begin{tabular}{|l|l|l|l|l|l|l|l|l|l|l|}
\hline
 Pretrained model /              & Emotion & Anger & Boredom & Calmness & Concentration & Determination & Excitement & Interest & Sadness & Tiredness \\
Acoustic    Features         & Regression & & & & & & & & & \\
               & model & & & & & & & & & \\
\hline
  wav2vec2.0 (Base)             & Architecture 1       &    0.328    &  0.466      &  0.509      &  0.459      &    0.442    &   0.382     &   0.310     &     0.394   &    0.468    \\
     & Architecture 2 &     0.339   &  0.47      &    0.520    &   0.457     &   0.462     & 0.394       &   0.329     &    0.394    &    0.468    \\
\hline
wav2vec2.0 (Large)    & Architecture 1 &   0.138 & 0.129 & 0.266 & 0.333 & 0.311 & 0.278 & 0.156 & 0.230 & 0.407     \\
               & Architecture 2     &   0.125 & 0.261 & 0.353 & 0.392 & 0.206 & 0.235 & 0.141 & 0.097 & 0.413  \\
\hline
HuBERT (Base)         & Architecture 1 & 0.418 & 0.545 & 0.566 & 0.526 & 0.526 & 0.441 & 0.403 & 0.478 & 0.544 \\
               & Architecture 2 & 0.441  & 0.552 & 0.567 & 0.526 & 0.532 & 0.455 & 0.415  & 0.485 & 0.546      \\
\hline
HuBERT (Large)   & Architecture 1 &  0.450 & 0.566 & 0.583 & 0.538 & 0.539 & \textbf{0.485} & \textbf{0.427} & 0.505 & \textbf{0.560} \\
               & Architecture 2     & \textbf{0.467}  & \textbf{0.575} & \textbf{0.587} & \textbf{0.542} & \textbf{0.547} & 0.476 & 0.426 & \textbf{0.518}  & 0.557 \\
\hline
MFCC Features & Architecture 1  &    0.231    &    0.358    &    0.404    &    0.290    &   0.363     &    0.301    &    0.148    &    0.125    &    0.390    \\
               & Architecture 2  & 0.180 & 0.343 & 0.379 & -0.017 & 0.327 & 0.295 & 0.162 & 0.195 & 0.342       \\
\hline
Baseline \cite{baseline}       &    & 0.428  & 0.545 & 0.559 & 0.524 & 0.531 & 0.453 & 0.431 & 0.476 & 0.55      \\
\hline
\end{tabular}
\end{adjustbox}
\caption{Summary of Spearman correlation ($\rho$) for Architecture 1: CNN+LSTM+FFNN, and Architecture 2: CNN+LSTM+Attention+FFNN with wav2vec2.0 and HuBERT foundation models as feature extractors. HuBERT-Large with Architecture 2 provides the \textbf{best} performance.}
\label{tab:all}
\end{table*}

\vspace{-2.5mm}
\subsection{Acoustic Features} 
In many emotion classification tasks, a standard set of speech features such as Mel spectrogram, Mel-Frequency Cepstral Coefficients (MFCC), and raw spectrogram are studied \cite{pandey2019deep}. Other non-semantic speech tasks like disfluency detection have also used MFCC and filter banks as their baseline method \cite{mohapatra2023efficient}. We have implemented acoustic feature extraction using a 40-dimensional mel-filterbank with cut-off frequencies at 0 Hz and 8000 Hz with a 25 ms window.
 
\vspace{-2.5mm}
\subsection{Emotion Regression Networks} \label{regressor}
After the embeddings have been extracted from the pre-trained models or using acoustic features, we use different types of regression networks to predict emotion share. In this subsection, we discuss different regression models. This part of the model is essential because the pre-trained self-supervised speech features are obtained from the models which are not specifically trained for emotion detection/share problems. For all the models discussed below, let us consider that the output from the pre-trained model (or acoustic features) is of dimension \textit{(N, L, W)} where \textit{N}, \textit{L}, and \textit{W} represents the batch size, sequence length, and feature size respectively. 

\noindent \textbf{Architecture 1 :} Architecture 1 comprises 1D CNN, an LSTM layer, and a feed-forward neural network (FFNN). 1D CNN layers increase the feature size from W to $W_c$ but help in summarising the time series data and reduce the sequence length to $\frac{L}{2}$. The output of the 1D CNN layers is fed into a 2-layer LSTM network to capture any time dependencies in the signal in the context of the emotion share task. The output of the LSTM layer is of the dimension \textit{(N, L, $W_l$)}. In the next step, we take the mean of the feature vector across all the time points of the LSTM output to obtain a tensor of dimension \textit{(N, $W_l$)}. For a sample in a batch with $L_i$ sequence length, the mean vector ($\overline{f_l}$) of size $W_l$ can be obtained using Eq.\ref{mean}. In Eq.\ref{mean}, $f_l^i$ represents the LSTM output vector at $i^{th}$ time step. The mean output is then fed into FFNN to get the emotion share. It is important to note that the emotion regression network is different for all the emotions and they are trained separately too. 
\vspace{-2.5mm}
\begin{equation}
    \overline{f_l} = \frac{\sum_{i=1}^{L_i} f_l^i}{L_i}
\label{mean}
\end{equation}

\noindent \textbf{Architecture 2 :} Architecture 2 is same as the Architecture 1 till the LSTM layer. In Architecture 2, after LSTM, we introduce the self-attention mechanism. Past works~\cite{emo_Attn} demonstrate that self-attention improves performance in emotion detection tasks. We calculate scaled dot-product attention~\cite{attn} using Query (\textit{Q}), Key (\textit{K}), and Value (\textit{V}). In our model, we consider \textit{K} and \textit{V} to be the same as the LSTM output. Query vector (\textit{Q}) is calculated from the final hidden state ($h_n$) of the LSTM. $h_n$ has the dimension of \textit{(2, N, $W_l$)} since we are using 2 hidden layers. To obtain Q from the $h_n$, we first transform $h_n$ to the dimension \textit{(N,2*$W_l$)}. Let us represent the transformed $h_n$ as $h_n'$. Q is obtained from $h_n'$ using a simple matrix multiplication shown in Eq.\ref{q}. In Eq.\ref{q}, the dimension of $W_Q$ is (2*$W_l$, $W_l$). In the next step, the attention (\textit{A}) is calculated using Eq.\ref{at}. Attention \textit{A} is then fed into FFNN to predict the emotion share. Fig. \ref{fig:Architecture_Diagram_}.(c) illustrates this architecture with pseudo-dimensions. 
\vspace{-2.5mm}
\begin{subequations}
	\begin{align}
		& Q = matmul(W_Q, h_n')   \label{q} \\
		& A = softmax\Biggl(\frac{QK^T}{W_l}\Biggl)V  \label{at} 
	\end{align}
\end{subequations}
\vspace{-3.5mm}
\subsection{Training Details}
We run the experiments on an Ubuntu OS server equipped with NVIDIA TITAN RTX GPUs with PyTorch framework. We use the Adam optimizer to train the model with mini-batch gradient descent. We use early stopping based on validation data for the final model selection. We have used 128 as the batch size with 1e-4 as the rate of learning. More details on our hyperparameter settings can be found in the released codebase\footnote{ \url{https://github.com/payalmohapatra/EmotionShare_ACMMM23.git}}.

\section{Experimental Results} \label{sec:results}
In this section, we discuss results obtained using embeddings from different pretrained models and different emotion regression networks discussed in the above sections. To assess the quality of fit we use Spearman correlation ($\rho$) \cite{spear} as it is used in the baseline paper \cite{baseline} too. Spearman correlation indicates how well can the relationship between the two variables be described using a monotonic function and it is a favored metric for ranking-based values. 
\vspace{-4.5mm}
\subsection{Evaluation on different Speech Embeddings}
Table \ref{tab:all} shows the $\rho$ value for all 9 emotions obtained from all the embeddings and acoustic features. Table \ref{tab:all} clearly shows that HuBERT-Large embeddings are the most effective. To understand the effect each embedding/acoustic feature has on the prediction, we show the percentage change in the average $\rho$ (among 9 emotions) value with respect to the baseline model \cite{baseline} in Table \ref{tab:pretrain}.  It can be clearly noted from Table \ref{tab:all} that the HuBERT-Large embeddings outperform all other embeddings as well as acoustic features. Also, HuBERT-Large embeddings outperform the baseline by 4.6\%. It is important to note that the \textbf{current task of predicting the emotion share is complex} as the ratings are dependent on human annotators. If we change the group of annotators, \textbf{the ground truth might vary}. Additionally, the data originates from \textbf{multilingual sources} but the pretrained models are trained only in the English language, making the task further complex. Therefore, for such a complex and uncertain task, \textbf{an improvement of 4.6\%} is significant. 

\begin{figure*}
  \centering
  \includegraphics[width=160mm, scale=0.7]{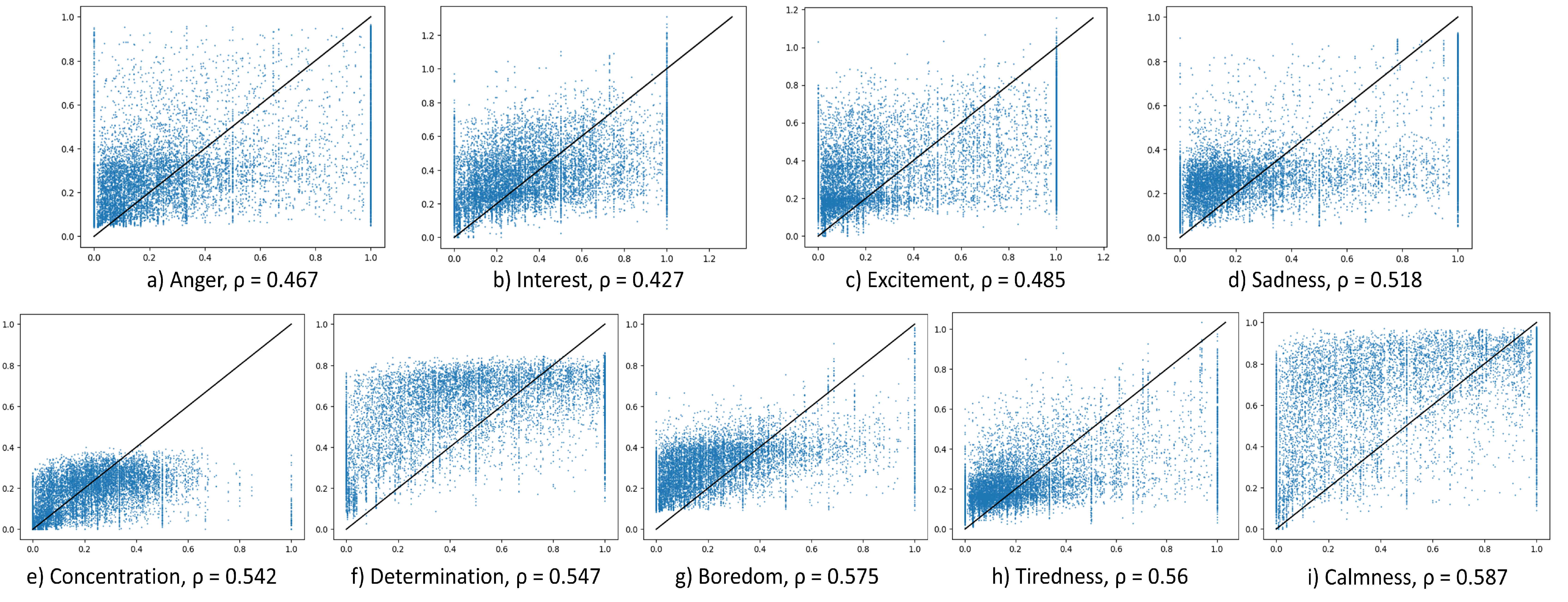}
  \caption{Scatter plot to show the quality of fit. The x-axis is the true value and the y-axis is the predicted value. The black line is a y=x line; points lying on it are the perfect prediction.}
 \label{fig:scatter}
\end{figure*}
% \vspace{-2 mm}

Although HuBERT-Large embedding performs the best, it does not indicate that using a large embedding size necessarily outperforms a smaller counterpart. This is evident from the fact that the wav2vec2.0-Base model outperforms the wav2vec2.0-Large model by 55\%. \textbf{HuBERT (base and Large) embeddings outperform all other models as it is able to capture more non-semantic elements in the audio due to their training method}. They inherently capture more abstract representations of a language which is possibly the reason for their better performance over wav2vec2.0.

Fig.\ref{fig:scatter} shows the scatter plot for all emotions on the \textit{dev} dataset. It reveals the regression imbalance in the data for most of the emotions. This can be one of the reasons for the correlation ($\rho$) to be low in our model as well as in the baseline model. The current performance can be further improved by adopting techniques to address the imbalance in regression data rather than feature extraction or updated architectures.
\vspace{-2.5mm}
\begin{table}[H]
\centering
\begin{adjustbox}{width=75mm,center}
\begin{tabular}{|c|c|c|c|}
\hline
Pretrained model / & Average &  Baseline Average & Percentage Change (\%) :\\
Acoustic Features &  $\rho$ (a) & $\rho$ (b) & $\frac{100*(a-b)}{b}$ \\ 
\hline
wav2vec2.0 (Base) & 0.425 &   & -15\\
wav2vec2.0 (Large) & 0.274 &  &  -45.2 \\
HuBERT (Base) & 0.502 & 0.500 & 0.4\\
HuBERT (Large) & 0.523 &  &  4.6 \\
Acoustic Features & 0.299 &  & -40.2 \\
\hline
\end{tabular}
\end{adjustbox}
\caption{Percentage change in average $\rho$ for different embeddings/acoustic features compared to the baseline on \textit{Dev} dataset. HuBERT-Large gives the most improvement.}
\label{tab:pretrain}
\end{table}
\vspace{-6.5mm}
\subsection{Evaluation on different Emotion Regressor networks}
\begin{figure}[htbp]
  % \centering
    \begin{adjustbox}{width=65mm,center}
      \includegraphics[scale=0.5]{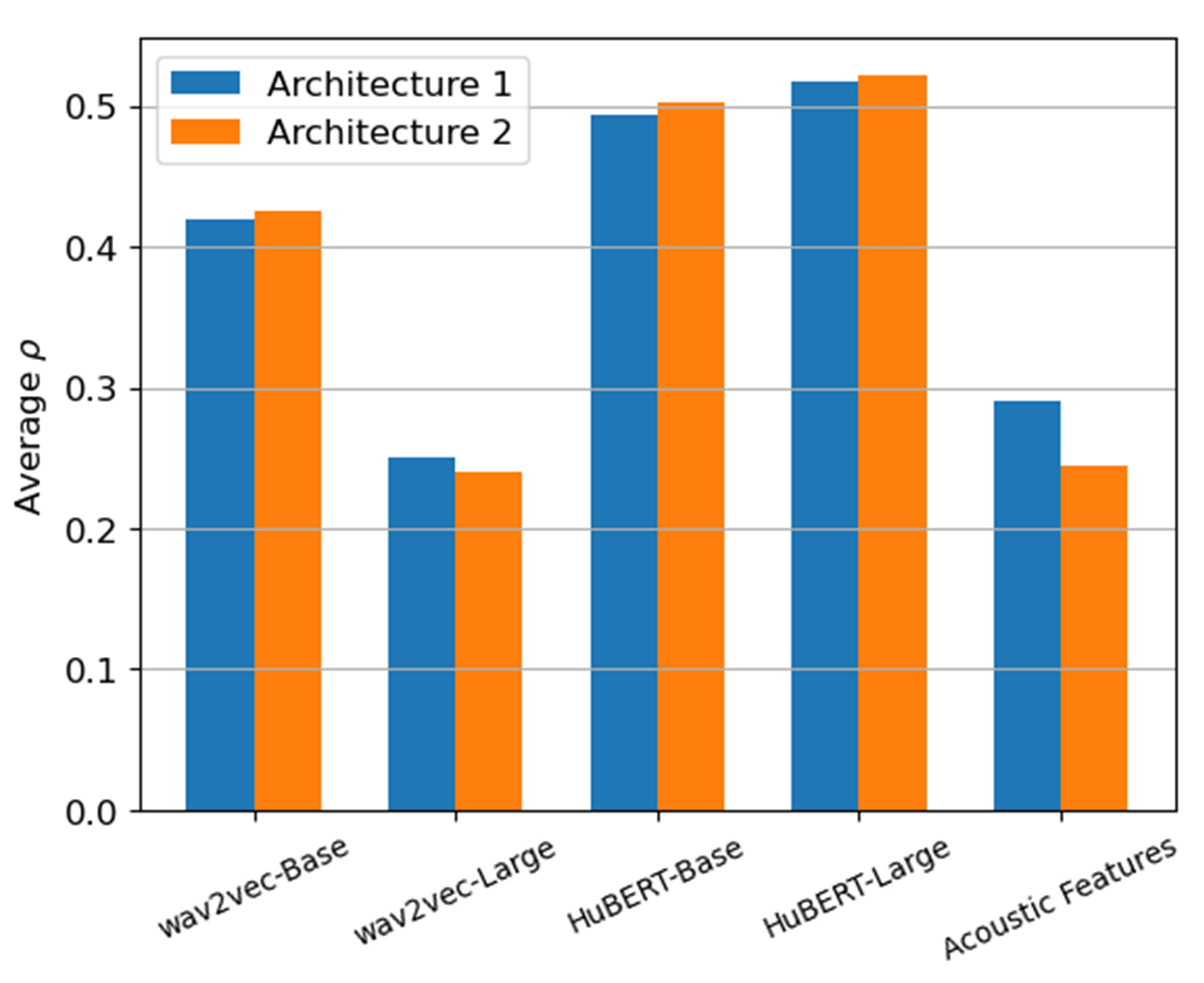}
    \end{adjustbox}
      \caption{Comparing the results from Architecture 1 and 2.}
 \label{fig:archit}
\end{figure}

In the overall model architecture, the emotion regressor network is the segment of the model which learns the paralinguistic part of the speech. Hence, it is extremely important to tune the network well. To that extent, as described in Sec.\ref{regressor}, we are using two types of emotion regressor networks. The only difference between Architecture 1 and 2 is the presence of a self-attention mechanism in Architecture 2. Additionally, we have also done some experiments with a multi-headed transformer network~\cite{attn}, which performed very poorly so we exclude it from further analyses. Fig.\ref{fig:archit} shows the average $\rho$ value for architecture 1 and 2 for different embeddings/acoustic features. It can be observed from Fig.\ref{fig:archit} that for wav2vec2.0-Base, HuBERT-base, and HuBERT-Large, Architecture 2 performs better by 1-1.8 \%. While for wav2vec2.0-Large and acoustic features, Architecture 2 performs worse. These observations signify that \textbf{attention can improve the model} but if the initial embeddings/features are not good then attention just hurts the model by increasing the parameters. 
\vspace{-1.5mm}
\vspace{-3mm}
\section{Conclusion}\label{sec:conclude}
In this work, we view speech emotion understanding as a perception by formulating a multi-label regression task rather than the commonly studied speech emotion recognition, a classification task. We demonstrate the effectiveness of our approaches on a unique dataset corpus with multilingual speakers to identify the share of people who perceive a given speech as one of the nine prominent emotions. We provide insights into using different foundation models as feature extractors and the role of attention in building an emotion regressor. Emotion relies on the non-semantic or paralinguistic aspects of speech. HuBERT-LARGE embeddings followed by a self-attention-based sequence model provide the best performance. In the Future, we are interested in addressing the current bottlenecks - 1) the use of foundation models pretrained only on English speech, 2) deep imbalance in the data for regression tasks, and 3) unavailability of meta-data like language identifiers, speaker characteristics(age, gender, etc.) as additional features.

% \section{Acknowledgement}
% We gratefully acknowledge funding from NSF awards 1834701 and 2038853.
% \todo{Before submitting change the following : \\
% * write wav2vec2 and HuBERT in times new roman or some other font \\
% * Write what we submitted and what new stuff we did. \\
% * Transformer results in text \\
% * Write about the training scheme --- which GPU and all \\
% * Make uniform wav2vec2 or wav2vec2.0\\
% * make uniform dev and val references}
\bibliographystyle{ACM-Reference-Format}
\balance
\bibliography{sample-base}

\end{document}